\documentclass[10pt,twocolumn,letterpaper]{article}

\usepackage{iccv}              % To produce the CAMERA-READY version

\usepackage{pifont}  
\usepackage{soul}
\usepackage{xcolor,cancel}
\usepackage[switch]{lineno}
\usepackage{tabularx}

\newcommand{\crossmark}{\ding{55}}

\definecolor{iccvblue}{rgb}{0.21,0.49,0.74}
\usepackage[pagebackref,breaklinks,colorlinks,allcolors=iccvblue]{hyperref}

\def\paperID{8} % *** Enter the Paper ID here
\def\confName{ICCV}
\def\confYear{2025}

\title{Efficient Depth- and Spatially-Varying Image Simulation for Defocus Deblur}

\author{Xinge Yang$^{1}$\thanks{Work done during Meta internship.} \quad Chuong Nguyen$^{2}$ \quad Wenbin Wang$^{2}$ \quad Kaizhang Kang$^{1}$ \\ Wolfgang Heidrich$^{1}$ \quad Xiaoxing Li$^{2}$\\ \\
KAUST$^{1}$ \quad Meta Reality Labs$^{2}$ %\\
}

\begin{document}
\maketitle
\begin{abstract}
Modern cameras with large apertures often suffer from a shallow depth of field, resulting in blurry images of objects outside the focal plane. This limitation is particularly problematic for fixed-focus cameras, such as those used in smart glasses, where adding autofocus mechanisms is challenging due to form factor and power constraints. Due to unmatched optical aberrations and defocus properties unique to each camera system, deep learning models trained on existing open-source datasets often face domain gaps and do not perform well in real-world settings. In this paper, we propose an efficient and scalable dataset synthesis approach that does not rely on fine-tuning with real-world data. Our method simultaneously models depth-dependent defocus and spatially varying optical aberrations, addressing both computational complexity and the scarcity of high-quality RGB-D datasets. Experimental results demonstrate that a network trained on our low resolution synthetic images generalizes effectively to high resolution (12MP) real-world images across diverse scenes.
\end{abstract}
\begin{figure}[t]  
    \centering  
    \includegraphics[width=\linewidth]{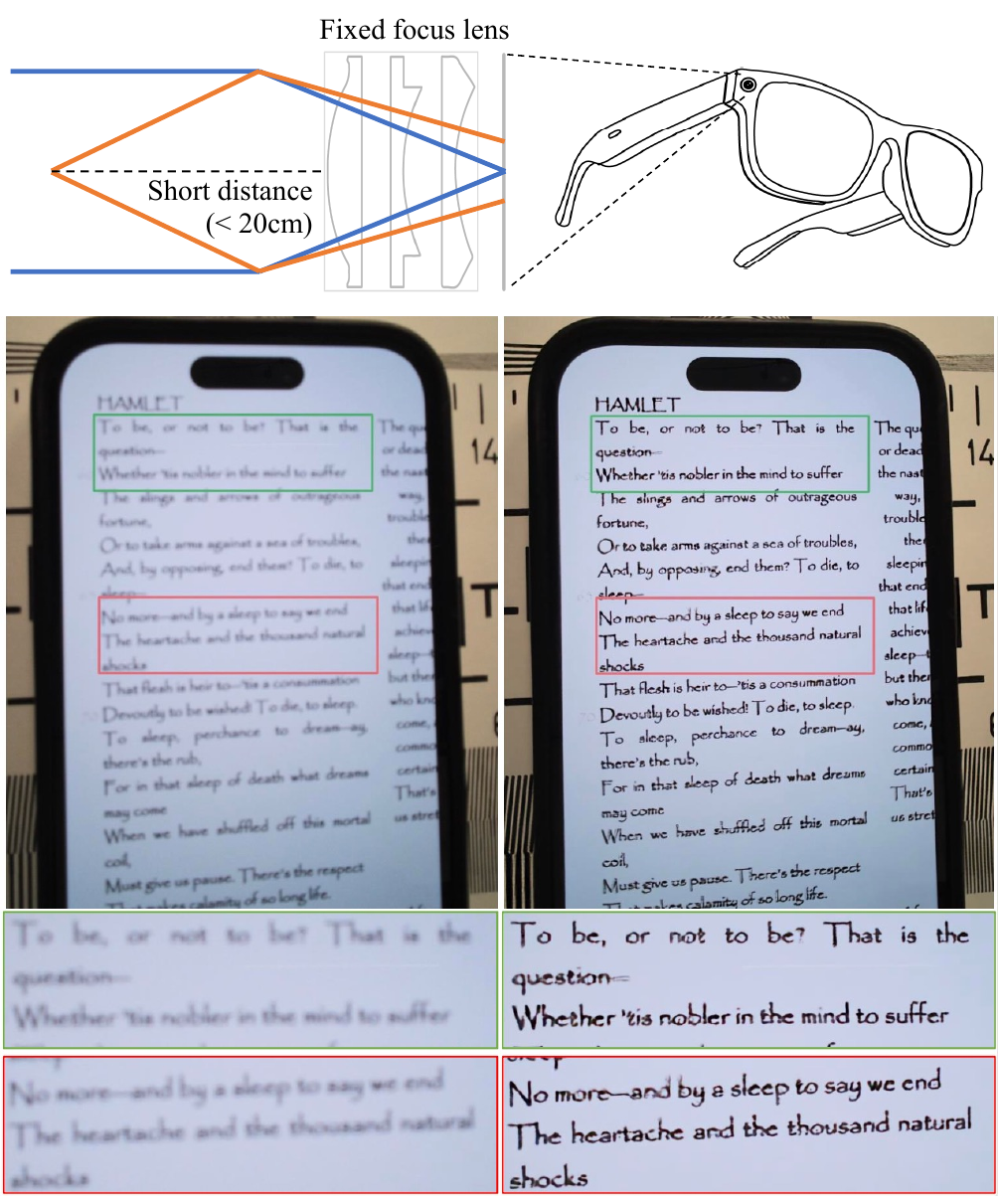}  
    \caption{\textbf{Motivation and real-world results for our proposed depth-varying dataset synthesis approach.} Large aperture fixed-focus lenses struggle to capture clear images at short distances (typically $<$20 cm), creating challenges for devices like smart glasses to perceive the physical world. Our efficient depth-varying dataset synthesis approach enhances computational photography algorithms for real-world defocus scenes. Bottom row compares the raw captured image (left) with our restored result (right), demonstrating promising results in defocus deblurring, optical aberration correction, and noise reduction.}  
    \label{fig:teaser}
\end{figure}

\section{Introduction}

The demand for computational photography algorithms to perform well in defocus scenarios is growing rapidly, as modern optical lenses often employ large apertures~\cite{Sheil_2019, Brady_2009, Blahnik_2021, Yang_2024Lens}. Large aperture sizes can reduce noise levels; however, they also result in a shallow depth of field, causing out-of-focus objects to appear blurry~\cite{Kuthirummal_2011}. In many cases, the defocus effect degrades the image quality and reduces the amount of information that can be extracted from the physical world, especially for edge-device cameras, for example, those on smart glasses.

Incorporating autofocus modules~\cite{Groen_1985,Kehtarnavaz_2003,ning2004auto} or novel extended depth-of-field optics~\cite{yang2024end,Yang_2024Lens,Sitzmann_2018} can ensure a wide focus range; however, these solutions are often constrained by factors such as device form factor, battery consumption, and immature manufacturing technologies. On the algorithmic side, both classical image deblurring methods~\cite{Zhuo_2011,Trouve_2011,D_Andres_2016,Liu_2020} and deep learning approaches~\cite{Xin_2021,Abuolaim_2020,Ma_2022,Quan_2024,ruan2023revisiting,Quan_2023,Ruan_2022,Yanny_2022} have been explored for defocus deblurring and restoration of image aberrations. Among these methods, neural networks typically yield higher image quality with fewer restoration artifacts. However, mismatches in optical aberrations, defocus scales, and noise statistics across different camera sensors prevent models trained on existing open-source datasets~\cite{Abuolaim_2020,Lee_2021,Ma_2022,Ruan_2021} from being directly applied to customized cameras, resulting in a significant domain gap.

To address the dataset gap, either new real-world datasets can be captured, or synthetic datasets can be employed. Capturing real-world datasets is generally expensive and time-consuming due to the requirement of covering varying spatial positions and depths. Synthetic dataset generation for defocus scenarios necessitates the simultaneous simulation of both depth-dependent defocus (caused by different pixel depths) and spatially-varying optical aberrations (caused by different pixel radial positions). However, existing optical simulation approaches either ignore depth-dependent defocus, considering only the focal plane~\cite{Chen_2021,Tseng_2021,Sun_2020}, or overlook spatially-varying optical aberrations~\cite{Wang_2021,Gur_2019}. There are two major challenges: First, simultaneously accounting for both optical spatial and depth variances renders the image simulation process computationally expensive~\cite{Yang_2024}. Second, there is a lack of high-quality, high-resolution RGBD datasets suitable for photorealistic dataset synthesis. These two challenges hinder large-scale dataset synthesis for dynamic real-world scenes. In summary, suitable training datasets for defocus deblur networks are crucial yet often lacking, especially when working with specific camera systems.

In this paper, we propose an efficient and scalable dataset synthesis approach for optical systems with depth- and spatially-varying effects. We first demonstrate (Sec.\ref{sec:results}) that synthetic datasets assuming planar depth perform poorly in real-world scenarios exhibiting significant defocus effects. Subsequently, we use a smart glasses fixed-focus camera as a test case to evaluate our proposed dataset synthesis approach. We comprehensively model depth-dependent defocus, spatially-varying optical aberrations, sensor quantization errors, and sensor noise. To address the limited availability of RGBD datasets, we apply DepthAnythingV2~\cite{yang2024depth} to high-quality RGB datasets and appropriately scale the estimated depth maps within our pipeline. Recognizing that spatial variance is minimal within small image patches from a 12-megapixel camera sensor, we disregard local spatial variance in low-resolution training batches. Instead, we incorporate positional encoding for each pixel to capture global spatial variance and encode ISO values to represent noise levels. This approach efficiently addresses the computational and dataset challenges inherent in defocus image simulation, enabling large-scale training data generation (Sec.~\ref{sec:method}).

Our experimental results show that a simple network trained on low-resolution synthetic images can deliver promising results on 12-megapixel full-resolution images across diverse real-world scenes (Sec.~\ref{sec:results}). The proposed approach is efficient as it supports fast on-the-fly training image synthesis, removes the need for point spread function (PSF) calibration or real-world image fine-tuning. With our proposed approach, a fixed-focus camera can image clearly for close objects, extending the usable scenarios for many applications. 
Building on this success, we demonstrate downstream applications for daily use cases with smart glasses, including short-distance optical character recognition (OCR), and 3D digital asset generation, where our proposed approach can greatly improve the final quality. In summary, our key contributions are as:
\begin{itemize}
\item We propose an efficient and scalable dataset synthesis approach that simultaneously models both spatially varying optical aberrations and depth-dependent defocus.
\item We address the lack of high-resolution RGBD datasets by applying pseudo depth maps, generated using state-of-the-art depth estimation models, to augment existing high-quality RGB datasets.
\item We establish an end-to-end training pipeline that effectively generalizes from low-resolution synthetic training data to high-resolution real-world images.
\end{itemize}
\section{Related works}

\subsection{Photorealistic Synthetic Dataset}
High-fidelity synthetic datasets with accurate physical modeling are effective to train networks that can generalize well to the real world~\cite{Brooks_2019, Chen_2021, Bhat_2021, Wei_2020}. Till now, research works have been done for noise models~\cite{Brooks_2019, Wei_2020}, spatially-varying optical aberrations~\cite{Chen_2021, Tseng_2021, Sun_2020}, with the promising pipeline unprocessing images from the sRGB space back to the RAW image space to simulate sensor noise~\cite{Brooks_2019, wei2021physics} and optical aberrations in the RAW domain~\cite{Chen_2021}.

However, current research works usually focus only on optical simulation at the focus plane, while ignoring defocus effects~\cite{Chen_2021,Tseng_2021,Sun_2020}, or modeling defocus effects while ignoring spatially varying optical aberrations~\cite{Wang_2021, Gur_2019,Wang_2023,Ruan_2024}. The challenge arises from the rapidly changing optical aberrations, including both variations across the imaging plane and variations with distance from the camera. Accurately modeling these effects not only requires a large degree of freedom to store the PSFs, but is also computationally expensive for high-resolution dataset generation, as each pixel has independent spatial positions and depths. The limited existing works~\cite{yang2024depth,Luo_2024} are not applicable for large-scale training dataset synthesis. In this work, we propose an efficient and accurate image simulation approach that considers both depth-dependent defocus effects and spatially varying aberrations. The proposed method not only greatly reduces the computational time required for synthetic training data generation but also maintains high fidelity in the simulated images.

\subsection{Single Image Defocus Deblur}
Image deblurring is a long-standing problem that aims to recover sharp and clear images from various types of blur, such as motion blur~\cite{Sun_2015,Su_2017,Kupyn_2019,Zhang_2019,Nah_2017}, defocus blur~\cite{Ruan_2024,Ruan_2021,Ma_2022,Quan_2023,Quan_2023imp,Lee_2021,ruan2023revisiting,Quan_2024}, and environment blur~\cite{Song_2023,Shi_2022}. Both classical methods~\cite{Zhuo_2011,Trouve_2011,D_Andres_2016,Liu_2020} and deep learning approaches~\cite{Xin_2021,Abuolaim_2020,Ma_2022,Quan_2024,ruan2023revisiting,Quan_2023,Ruan_2022} have been explored and have shown promising results. Typically, large amounts of high-quality data are required to train deep networks effectively. The existing defocus deblurring dataset~\cite{Abuolaim_2020,Lee_2021,Ma_2022,Ruan_2021} is quite limited. Additionally, considering that the defocus characteristics and optical aberrations vary across different lenses, a network model trained on the open-source dataset often cannot be directly applied to images captured with another lens. Capturing enough high-quality training data for each lens is impractical and time-consuming. In this work, we propose a synthetic dataset generation approach that allows machine learning engineers to train image deblurring networks directly on synthetic data, with promising performance on real captured images.
\begin{figure*}[ht]
    \centering
    \includegraphics[width=\linewidth]{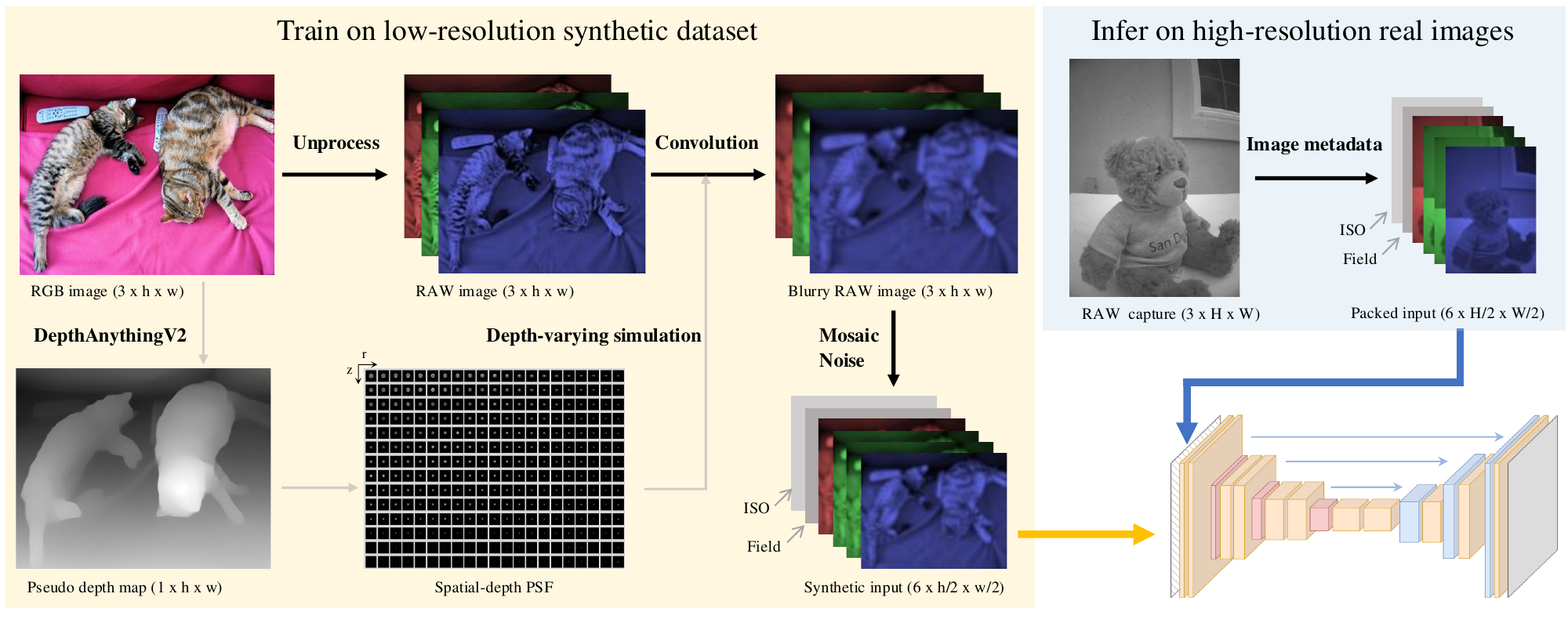}
    \caption{\textbf{Training and inference pipeline of the proposed approach.} Left: Images are unprocessed from RGB space to RAW space to simulate defocus blur, optical aberrations, sensor quantization, and noise. A pseudo depth map is predicted using the pretrained DepthAnythingV2~\cite{yang2024depth} model, then randomly scaled and utilized in the depth-varying defocus and spatially-vary aberration simulation. Noise signal at a random ISO level is added to the blurry RAW image. The image data, ISO channel, and radial field map are then packaged as network inputs. Top Right: During the inference stage on real-world images, the ISO value is read from photograph metadata, and the field map is computed on full-resolution images. Bottom Right: Instead of relying on complicated network architectures, a simple network (NAFNet~\cite{Chen_2022}) is adopted for image reconstruction.}
    \label{fig:pipeline}
\end{figure*}

\begin{figure}[ht]
    \centering
    \includegraphics[width=\linewidth]{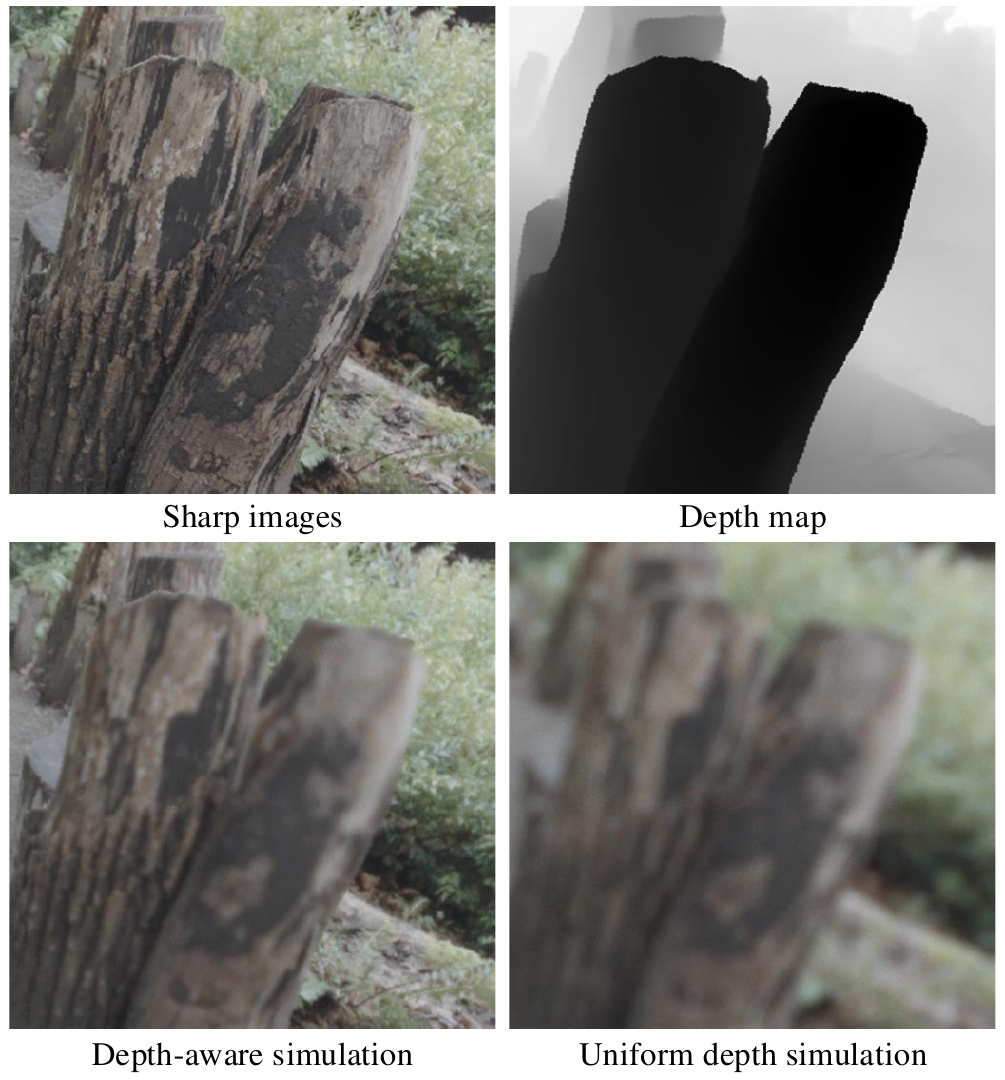}
    \caption{\textbf{Comparison of synthetic training image with and without depth-varying simulation.} Incorporating depth-varying defocus allows for more realistic simulations, reflecting real-world scenarios where objects at varying distances from the camera exhibit different levels of defocus.}
    \label{fig:render}
\end{figure}

\section{Methods}
\label{sec:method}

We first discuss our efficient depth-varying defocus and spatially-varying aberrated data generation pipeline in Sec.~\ref{sec:datageneration}, as well as illustrated in Fig.~\ref{fig:pipeline}. In Sec.~\ref{sec:implementation}, we discuss the dataset preparation and more implementation details 

\subsection{Efficient Defocus and Aberration Simulation}
\label{sec:datageneration}
To generate realistic synthetic images, we unprocess RGB images to RAW signal space by inverting the image signal processing (ISP) pipeline. This is based on two main considerations: first, the PSF of the lens is typically defined in the radiance space, while the camera ISP will apply nonlinear processing to the RAW signals~\cite{Chen_2021}. Second, the sensor noise, which greatly affects the image reconstruction results, also suffers from postprocessing~\cite{Brooks_2019}. The unprocessing and PSF convolution pipeline can be expressed as
\begin{equation}
    \mathbf{I}^{'} = \mathbf{P} * \mathcal{F}^{-1}(\mathbf{I}),
\end{equation}
where $\mathbf{I}$ is the input image, $\mathcal{F}^{-1}$ represents the ``unprocess'' as described in \cite{Brooks_2019}, $\mathbf{P}$ represents the PSF function and $*$ denotes the convolution operation. $\mathbf{I}^{'}$ is the blurred signal in the RAW signal space. Note that the PSF of the camera changes across both the image plane and different depths, which means each image pixel has independent imaging characteristics.

For an accurate image simulation, ideally, we have to perform convolution between each pixel with spatially-varying depth-dependent PSF $\mathbf{P}(u, v, z_{[u, v]})$, which can be expressed as
\begin{equation}
    \mathbf{I}^{'}_{[u, v]} = \mathbf{P}(u, v, z_{[u, v]}) * \mathcal{F}^{-1}(\mathbf{I})_{[u, v]} \label{eq:per_pixel},
\end{equation}
where $u$ and $v$ denote the normalized spatial position of the pixel on the image plane, and $z$ denotes the corresponding depth value. However, both the storage of per-pixel PSF and the per-pixel convolution are computationally expensive. Although some recent works successfully speed up the PSF representation problem~\cite{Denis_2015, Yanny_2020, Yang_2024}, the per-pixel convolution computation is still a time-consuming problem, particularly for large-scale dataset generation.

Based on two observations that (1) neural networks process high-resolution images by smaller patches, and (2) in a small image patch of a high resolution sensor, in-plane spatial positions ($u$, $v$) between different pixels do not vary too much, we believe \textbf{it is reasonable to ignore the in-plane spatial variance during the network training stage}. This simplification allows us to only focus on the depth variance of pixels, which significatnly improve the image simulation efficiency. Consequently, we simplify Eq.\eqref{eq:per_pixel} to
\begin{equation}
    \mathbf{I}^{'}_{[u, v]} = \mathbf{P}(z_{[u, v]}) * \mathcal{F}^{-1}(\mathbf{I})_{[u, v]} \label{eq:small_patch}.
\end{equation}

For a PSF at an unknown depth, it can be linearly interpolated by its neighbour as long as the sampling is dense enough. Also, the convolution operation is a linear operation, we can further rewrite Eq.\eqref{eq:small_patch} as
\begin{align}  
    \mathbf{I}^{'}_{[u, v]} &\approx (\sum_{z \in \mathbf{Z}} \alpha_{[u, v]} \mathbf{P}_{z}) * \mathcal{F}^{-1}(\mathbf{I})_{[u, v]} \label{eq:interp_psf}\\
                    &\approx \sum_{z \in \mathbf{Z}} \alpha_{[u, v]} (\mathbf{P}_{z} * \mathcal{F}^{-1}(\mathbf{I}))_{[u, v]} \label{eq:interp_img},
\end{align}
where $\mathbf{Z}$ is a discrete set of predefined depths, and $\alpha$ denotes the weight of interpolation. Eq.\eqref{eq:interp_psf} denotes the interpolation in the PSF space. However, a costly per-pixel convolution is still required. To further simplify computation,  Eq.\eqref{eq:interp_img} converts the problem to first compute convolution between base PSF functions at different depth $z\in\mathbf{Z}$ and input images, then interpolate in the image space. This greatly reduces the computation resources as matrix production and image convolution can be sped up with modern computation algorithms and hardware. Fig.~\ref{fig:render} compares synthetic images with and without depth-dependent PSF.

We follow the read and shot noise model as described in~\cite{Brooks_2019} to simulate the sensor raw. Here, $\lambda_{read}, \lambda_{shot}$ are the read and shot noise variance that depends on both analogue and digital gain. These two gain levels are set as a direct function of the ISO light sensitivity level, chosen manually by the user or automatically by the camera. The ISO can be read from the metadata.
\begin{equation}
	\label{eq:noise-model}
	\mathbf{I}^{''} \sim {\mathcal {N}}(\mu=\mathbf{I}^{'},\sigma ^{2}=\lambda_{read} + \lambda_{shot}\mathbf{I}^{'})
\end{equation}
Finally, consider a $b$-bit sensor (typically $b=10$), the $b$-bit $\mathbf{I}^{'''}$ can then be computed from $\mathbf{I}^{''}$ as follows:
\begin{equation}
	\label{eq:quantization}
	\mathbf{I}^{'''} = \lfloor{f_b(\mathbf{I}^{''})}\rceil
\end{equation}
where  $\lfloor{.}\rceil$ is \textit{rounding} to the nearest integer operator and $f_b(\mathbf{y}) = \text{min}(\mathbf{y}, 2^b-1)$ is a $b$-bit clipping function.

The network $\mathcal{U}$ is trained to reconstruct clear RAW images $\mathcal{F}^{-1}(\mathbf{I})$ from blurry and noisy RAW inputs $\mathbf{I}^{'''}$. The loss function is defined in the RGB space to prioritize the quality of the final RGB images. Specifically, the loss function is formulated as  
\begin{equation}  
    \mathcal{L} = \mathcal{L} \left( \mathcal{F'} \left( \mathcal{F}^{-1}(\mathbf{I}) \right), \mathcal{F'} (\mathcal{U}( \mathbf{I}^{'''}) \right),  
\end{equation}
where $\mathcal{F'}$ denotes the ISP, which can differ from $\mathcal{F}$ used in the unprocessing stage. $\mathcal{U}(\mathbf{I}^{'''})$ is the network output, given the noisy RAW $\mathbf{I}^{'''}$ as input. Notably, we set the gamma parameter in $\mathcal{F'}$ to 2.0 to emphasize dark regions in the reconstruction results. The loss function $\mathcal{L}$ comprises both pixel loss ($L_1$) and perceptual loss (LPIPS~\cite{Zhang_2018}).  

\subsection{Scalable Dataset Preparation}
\label{sec:implementation}

For synthetic training data generation, we use RGB images from the Adobe5k dataset~\cite{fivek} with the unprocessing manner to obtain simulated RAW captures~\cite{Brooks_2019}. This RGB to RAW space unprocessing is of great importance especially when we want to generate dataset in specific scenarios, for example close distance optical character recognition (OCR). 

\begin{table*}[t]
  \centering  
  \caption{\textbf{Quantitative evaluation on synthetic datasets across different synthetic dataset generation approaches.} The best performance for PSNR, SSIM, and LPIPS is highlighted. For each metric, $\uparrow$/$\downarrow$ indicates that higher/lower values are better, respectively. The results demonstrate that both depth-varying simulation and the incorporation of auxiliary channels in the network input enhance image reconstruction performance.}  
  % \begin{tabular}{@{}lcccccc@{}}
  \begin{tabularx}{\textwidth}{@{} >{\raggedright\arraybackslash}X cccccc @{}}  
    \hline
    Method & Training Dataset & Defocus & Input Data & PSNR $\uparrow$ & SSIM $\uparrow$ & LPIPS $\downarrow$ \\
    \hline
    PolyBlur~\cite{Delbracio_2021} & \crossmark & \crossmark & RGB & 26.18 & 0.7205 & 0.3235 \\
    LaDKNet~\cite{ruan2023revisiting} & Captured & \checkmark (Unmatched) & RGB & 25.90 & 0.7165 & 0.4441 \\
    Chen et al.~\cite{Chen_2022} & Synthetic & \crossmark & RAW-ISO-Field & 27.14 & 0.8196 & 0.2469 \\
    \hline
    Ablation \#1 & Synthetic & \checkmark & RAW & 27.60 & 0.8223 & 0.2233 \\  
    Ablation \#2 & Synthetic & \checkmark & RAW-ISO & 28.73 & 0.8589 & 0.1960 \\
    Ablation \#3 & Synthetic & \checkmark & RAW-Field & 28.14 & 0.8317 & 0.2013 \\
    Ours & Synthetic & \checkmark & RAW-ISO-Field & \textbf{30.03} & \textbf{0.8808} & \textbf{0.1553} \\
    \hline
    % \end{tabular}
  \end{tabularx}  
  \label{tab:performance}  
\end{table*}

\begin{figure*}[ht]
    \centering
    \includegraphics[width=\linewidth]{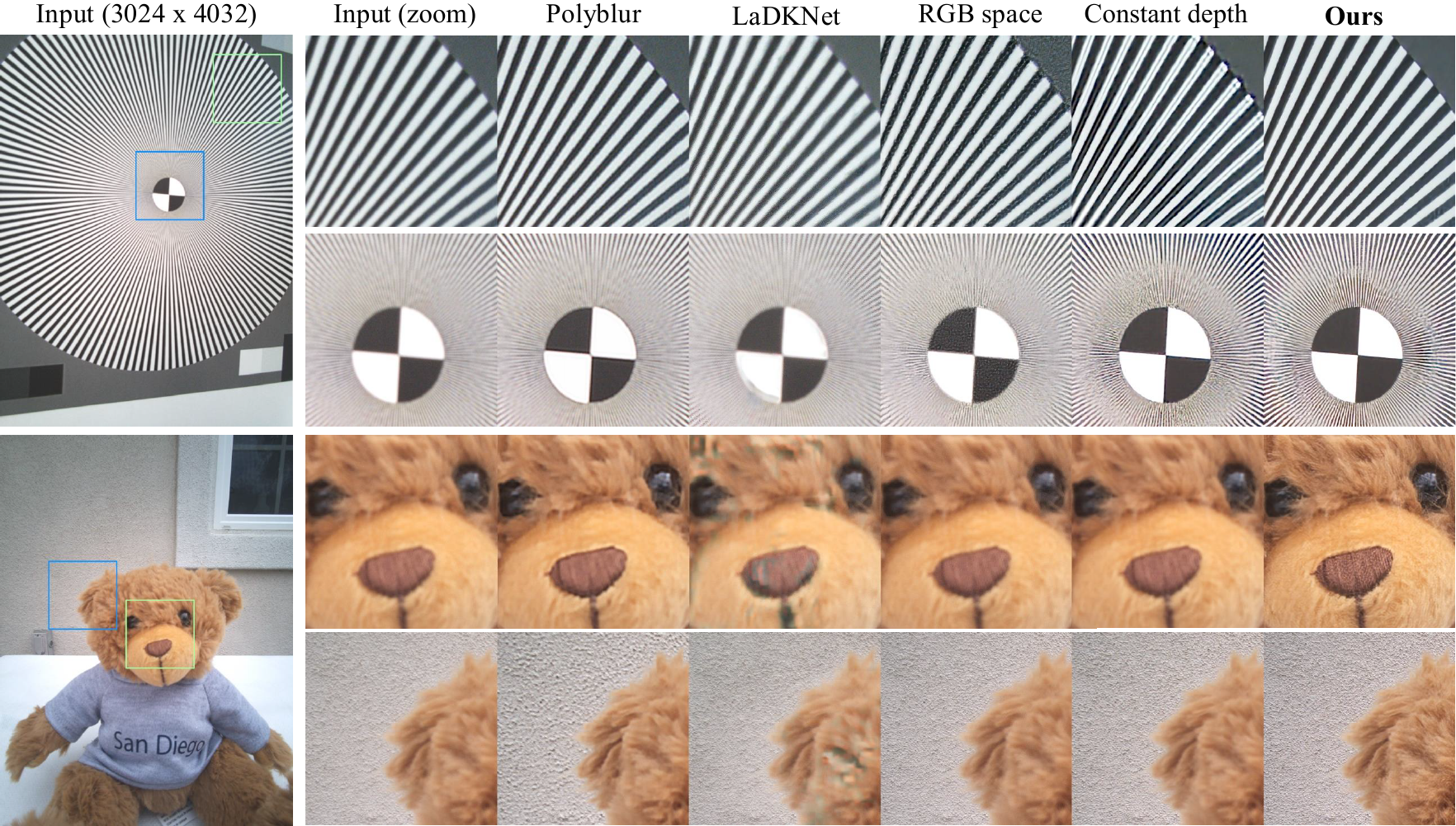}
    \caption{\textbf{Qualitative evaluation on 12MP real-world images with different defocus deblur methods and synthetic dataset generation.} From left to right: the classical deblurring algorithm (``Polyblur'')\cite{Delbracio_2021} exhibits limited capability, failing to produce high-quality images. The network (``LaDKNet'')\cite{ruan2023revisiting}, trained on an open dataset~\cite{Abuolaim_2020}, cannot be directly applied to our camera captures and produces artifacts due to inconsistent optics and sensor noise. Simulation in RGB image space results in images with structural artifacts due to inaccurate noise modeling. A network trained on synthetic datasets without depth-varying image simulation fails to deblur effectively, and also struggles with scenes that have varying depth ranges. For example, in the second case, the network is confused by the sharp wall and therefore fails to recover the teddy bear. Our proposed synthetic dataset generation helps the network implicitly learn to detect sharp and blurry regions, successfully recovering clear details such as the face of the teddy bear.}
    \label{fig:compare}
\end{figure*}

To address the lack of large-scale high-resolution RGBD datasets for close-up scenes, we employ the DepthAnythingV2~\cite{yang2024depth} model for depth estimation from RGB images before the training stage. Since the state-of-the-art depth estimation models only give relative depth maps, we scale them to absolute depths within our target depth range with multiple random scaling strategies, including linear, quadratic, and exponential functions.

Besides the augmentation coming from the random depth scaling, data augmentation is also applied at multiple stages, including geometric and pixel augmentation of RGBD images, unprocessing from RGB to RAW images, PSF augmentation, and post-processing from RAW to RGB images. Particularly, for PSF augmentation, we randomly apply a Gaussian blur with a small standard deviation to the PSFs to simulate the lens manufacturing and assembly errors in the real world.

Experiments are conducted using Meta Ray-Ban smart glass camera with a fixed focus distance set to infinity. The camera lens has a large aperture size (f-number 2.2), causing objects at short distances to appear significantly blurry due to defocus effects. PSFs at different spatial positions are computed using ZEMAX~\cite{ZemaxOpticStudio2023} with internal lens data, encompassing 20 depth stops from 10~cm to infinity and 20 radial stops from the sensor origin to the maximum field-of-view (FoV). Noise statistics for the camera sensor are calibrated as discussed in existing literature~\cite{Healey_1994, Brooks_2019} and supplementary materials.

\subsection{Tunable Auxiliary Channels} 

We employee auxiliary channels to tackle the blind deblur problem, including a single-channel ISO and a single-channel normalized radial position map stacked with the input image~(Fig.~\ref{fig:pipeline}). For the ISO channel, we employ a uniform map with a constant ISO value, which is scaled by 0.001 to match the range of the input data. The use of auxiliary channels serves two primary purposes. First, image noise statistics are highly dependent on the ISO value, and blur profiles within an image patch are significantly affected by radial position. Without explicit ISO information during training and inference, the network must independently infer noise levels, increasing its complexity and potentially leading to averaged outputs. Including the radial position channel enables the network to perform patch-specific deblurring for different regions, thereby simplifying its task. Second, during inference, the auxiliary channels can be adjusted to achieve the desired subjective image quality. For instance, doubling the ISO value enhances denoising to produce smoother images, while halving it reduces denoising to better preserve texture details.

We select the NAFNet~\cite{Chen_2022} for image reconstruction due to its efficiency and low latency. Training was performed using 256$\times$256 Bayer RGGB images comprising 6 input channels (4 RGGB Bayer patterns and 2 auxiliary channels). The network was trained on 8 H100 GPUs with a batch size of 64 using the AdamW optimizer, initialized with a learning rate of $10^{-4}$. Given the variability in spatial locations, gains, and depth scaling strategies, the training process was designed to cover a wide range of scenarios, thereby necessitating an extended training duration. In our experiments, training the network for 500 epochs takes approximately two days.
\section{Results}
\label{sec:results}
We assess the effectiveness of our dataset synthesis approach by comparing it with various alternative methods. Specifically, we evaluate different dataset synthesis techniques, as well as state-of-the-art deep learning models trained on open-source defocus deblurring datasets.

\subsection{Inference on Real-World 12-megapixel Images}
After training network with our proposed synthetic dataset, we directly use the network for inference on 12-megapixel real-world images captured across various scenes. The RAW captured data with the camera metadata are used as the network input, and the output RAW data undergoes a minimum post-processing for visualization.

In our experiments, we compare the defocus deblurring results obtained using a classical method (``Polyblur'')~\cite{Delbracio_2021}, the state-of-the-art deep learning method (``LaDKNet'')~\cite{ruan2023revisiting} with pretrained weights on the DPDD dataset~\cite{Abuolaim_2020}, and a NAFNet network~\cite{Chen_2022} trained with different synthetic dataset generation approaches. Several synthetic dataset generation approaches are chosen for comparison. In existing works, the most commonly used method is to directly apply PSF convolution to RGB images. Recent work by Chen et al.~\cite{Chen_2021} demonstrates promising results in removing spatially-varying optical aberrations using only synthetic datasets; however, they consider only the spatial variance of optical aberrations within the focus plane and ignore the defocus effect. To conduct a fair comparison with their method, we randomly assign constant depth maps to the input RGB images during image simulation, with a higher probability assigned to closer depths. By comparing with them, we want to prove the importance of depth-varying optical simulation.

In Fig.~\ref{fig:compare}, we show two example images reconstructed by different approaches. From the results, we observe that classical deblurring methods can slightly sharpen the images, but the improvement is quite marginal, and artifacts such as halos are introduced. Pretrained networks on open datasets cannot be generalized to our camera because the optical properties and noise statistics are different. Synthesizing data in the RGB image space leads to reconstructions with artifact noise patterns, especially in low-frequency regions such as the resolution chart. This issue arises because noise signals are not accurately modeled when simulating in the RGB image space, causing the network to fail to recover clear latent signals. For synthetic datasets without depth-varying optical simulation, the chart image is effectively deblurred, but the bear image remains blurry. This likely occurs because the network, trained with a constant defocus level across all pixels, is confused by sharp regions. For instance, the sharp background wall in the second example might lead the network to assume the entire image is in focus, preventing it from deblurring the bear's face. In short, training with constant depth maps makes the network unable to distinguish spatially varying defocus (see Supplement for more examples).

In contrast, the network trained with our proposed synthetic dataset generation approach performs much better. In the resolution chart image, both high-frequency and low-frequency regions are well reconstructed from blurry and noisy inputs without producing reconstruction artifacts. In the toy bear example, the face of the toy is successfully recovered, along with the text on the label in another spatial location. We believe this is because \textbf{with depth-varying optical simulation, the network model implicitly learns defocus detection from the spatially varying blurry inputs}, even though no explicit depth information is given, which explains the success and generality in the real world. Additional comparison examples on real-world captured images can be found in the Supplement.

% =====================================================================
% =====================================================================

\subsection{Quantitative Evaluation on Synthetic Dataset}
To quantitatively assess the effectiveness of our proposed method, we established a synthetic validation dataset that fully simulates spatially varying and depth-dependent optical aberrations, defocus, sensor noise at various ISO levels, and sensor quantitative errors. Specifically, we selected 2,000 RGB images from the EBB!~\cite{Ignatov_2020} dataset, each containing both close-up subjects and background scenes. The images were center-cropped and downsampled to a 512×512 resolution. Starting with all-in-focus images (with both subject and background in sharp focus), we applied depth estimation, random depth scaling, unprocessing, pixel-varying PSF convolution, and noise injection. For pixel-wise PSF calculation, the given PSF and the estimated depth map are used to interpolate the PSF for each image pixel~\cite{Yanny_2020}. For pixel-wise PSF convolution, we adopt the folding calculation method~\cite{yang2024depth}. Notably, the validation data incorporates spatial variance within a local image patch, which better reflects the real world and enables a more realistic evaluation of different methods.

Presented in the Table~\ref{tab:performance}, we compare defocus deblur results between a classical method (PolyBlur~\cite{Delbracio_2021, Eboli_2022}), the state-of-the-art defocus deblur model (LaDKNet~\cite{ruan2023revisiting}) trained on open source datasets (DPDD~\cite{Abuolaim_2020}), networks trained without considering varying depth maps~\cite{Chen_2021}, and networks trained with our depth-varying synthetic dataset. Experimental results demonstrate that all comparison methods cannot achieve satisfying results. Specifically, classical deblur methods can slightly sharpen the images, while the overall image quality is not satisfying. For LaDKNet, the domain gap between the open source training dataset and the target camera characteristics prevents it from functioning well on a new camera system. Synthetic datasets without depth-varying defocus modeling perform much worse, which we believe is because the network model fails to learn defocusing deblur capabilities from constant depth and invariant defocus maps. In contrast, with our proposed depth-varying dataset synthesis, we observe significant improvements in image restoration quality, demonstrating the effectiveness of our proposed approach.

We perform ablation studies on the same network (NAFNet) trained and evaluated using various auxiliary input channels (spatial position and camera ISO), as shown in Table~\ref{tab:performance}. With camera RAW capture inputs and no auxiliary channels, the reconstructed image quality surpasses previous results but falls short compared to using either the ISO or spatial position (``Field'') channels. This likely occurs because the model lacks information about the image patch location and noise level, leading it to generalize across all possible scenarios. When both ISO and spatial position channels are provided, the image quality improves significantly.

\begin{table}[t]
  \centering
  \caption{\textbf{Efficiency and performance comparison with full optical simulation}. Our proposed synthetic data generation approach significantly reduces rendering time and peak GPU memory consumption, enabling more efficient large-scale data generation while maintaining comparable or better performance.}
  \resizebox{\linewidth}{!}{  
    \begin{tabular}{@{}lcc@{}}  
      \hline
      & Full simulation~\cite{Yang_2024} & Ours \\
       \hline
      Rendering Time (ms) & 1306.9 & \textbf{16.3} \\
      Peak GPU Memory (GB) & 10.6 & \textbf{1.1} \\
      \hline
      \small PSNR (dB) & \textbf{30.09} & 30.03 \\
      \small SSIM & 0.8633 & \textbf{0.8808} \\
      \small LPIPS & 0.1849 & \textbf{0.1553} \\
      \hline
    \end{tabular}  
  }  
  \label{tab:efficiency}  
\end{table}

We further evaluated the efficiency and performance of our approach against full optical simulation (Table~\ref{tab:efficiency}). Our comparison encompassed per-pixel PSF convolution for spatially-varying optical aberrations following~\cite{Yang_2024}, depth-varying defocus, and sensor noise. The full optical simulation proves substantially more resource-intensive, requiring $\sim$80 times longer rendering time per training batch and consuming $\sim$9.6 times more peak GPU memory. These measurements were taken using (1, 3, 512, 512) image batches on a single A100 GPU. Despite these computational differences, our efficient training approach achieves comparable image quality as measured by PSNR, SSIM, and LPIPS metrics (Table~\ref{tab:efficiency}). These results confirm that \textbf{neglecting spatial variance within low-resolution training images for high-resolution camera sensors is a reasonable simplification while maintaining promising final performance}. Our efficient simulation approach significantly increases maximum batch size and reduces training time, accelerating algorithm development and offering broader benefits for computational photography algorithms.
\section{Applications}

Based on the success of the proposed approach, we explored two downstream applications: short-distance OCR and small object 3D reconstruction. Due to the small physical scale and fine details of text characters and small objects, these applications present significant challenges in balancing image quality and details for existing fixed-focus cameras in smart glasses, primarily because of their shallow depth of field.

\subsection{OCR}

\begin{figure}[t]
    \centering
    \includegraphics[width=0.95\linewidth]{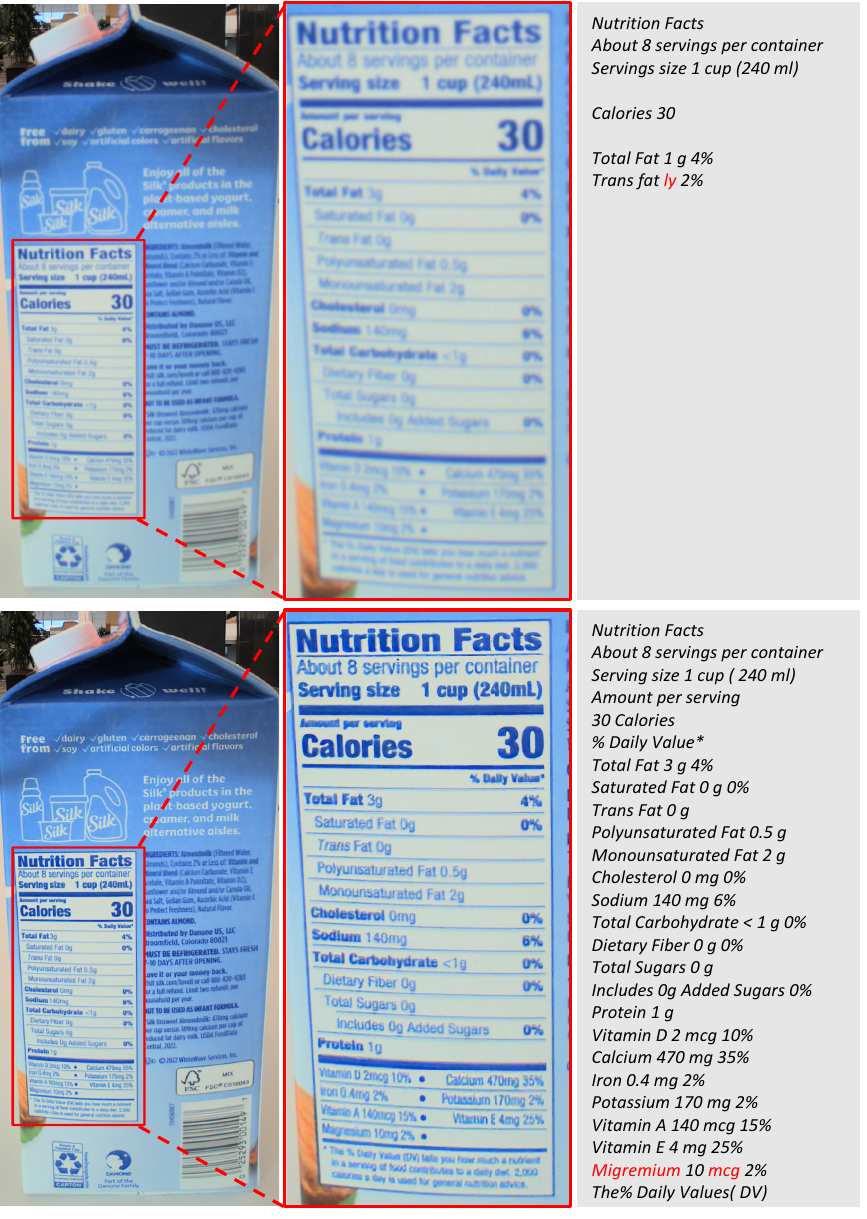}
    \caption{\textbf{Performance improvement in OCR for scene understanding.} Given an input image with texts captured at close distance (left), our depth of field extension successfully recovers details (right). Our result significantly improves OCR performance for both accuracy and detection rate. OCR results are generated with online program~\cite{wei2024general}, with errors marked in red.}
    \label{fig:ocr}
\end{figure}

Smart glass utilizes OCR technology to enhance user experiences by providing real-time text recognition and interaction capabilities. The OCR technology for smart glasses focuses on text captured from the user's point of view while wearing the glasses. This approach allows for seamless interaction with text in various environments, such as translating menus, adding business card contacts, or creating shopping lists. Image blurriness due to fixed focus can be challenging to text recognition models, particularly when capturing at a close distance. Fig.~\ref{fig:ocr} shows the OCR accuracy improvement using our approach.

\subsection{3D digital asset generation}

\begin{figure}[t]
    \centering
    \includegraphics[width=0.48\linewidth]{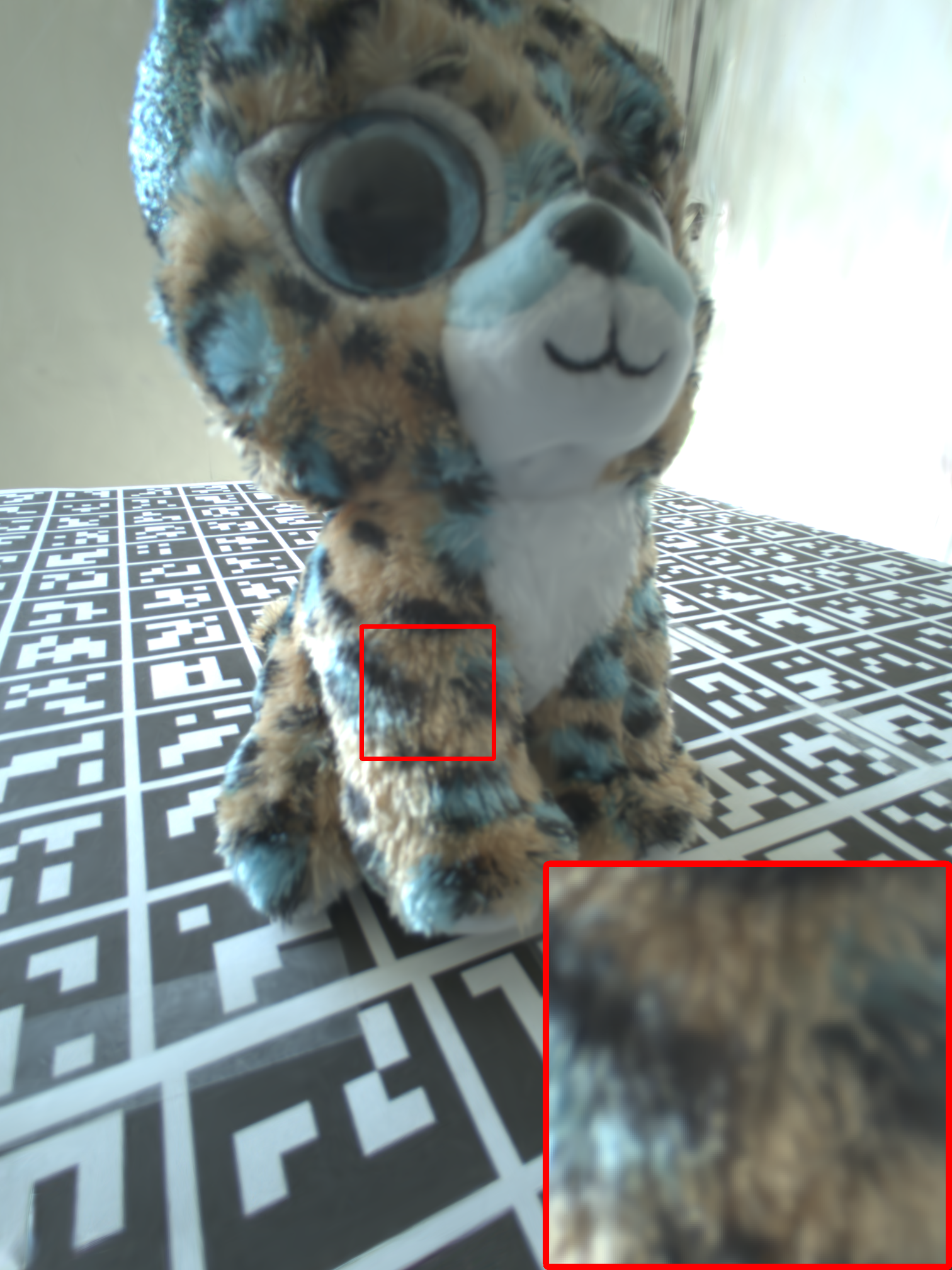}
    \includegraphics[width=0.48\linewidth]{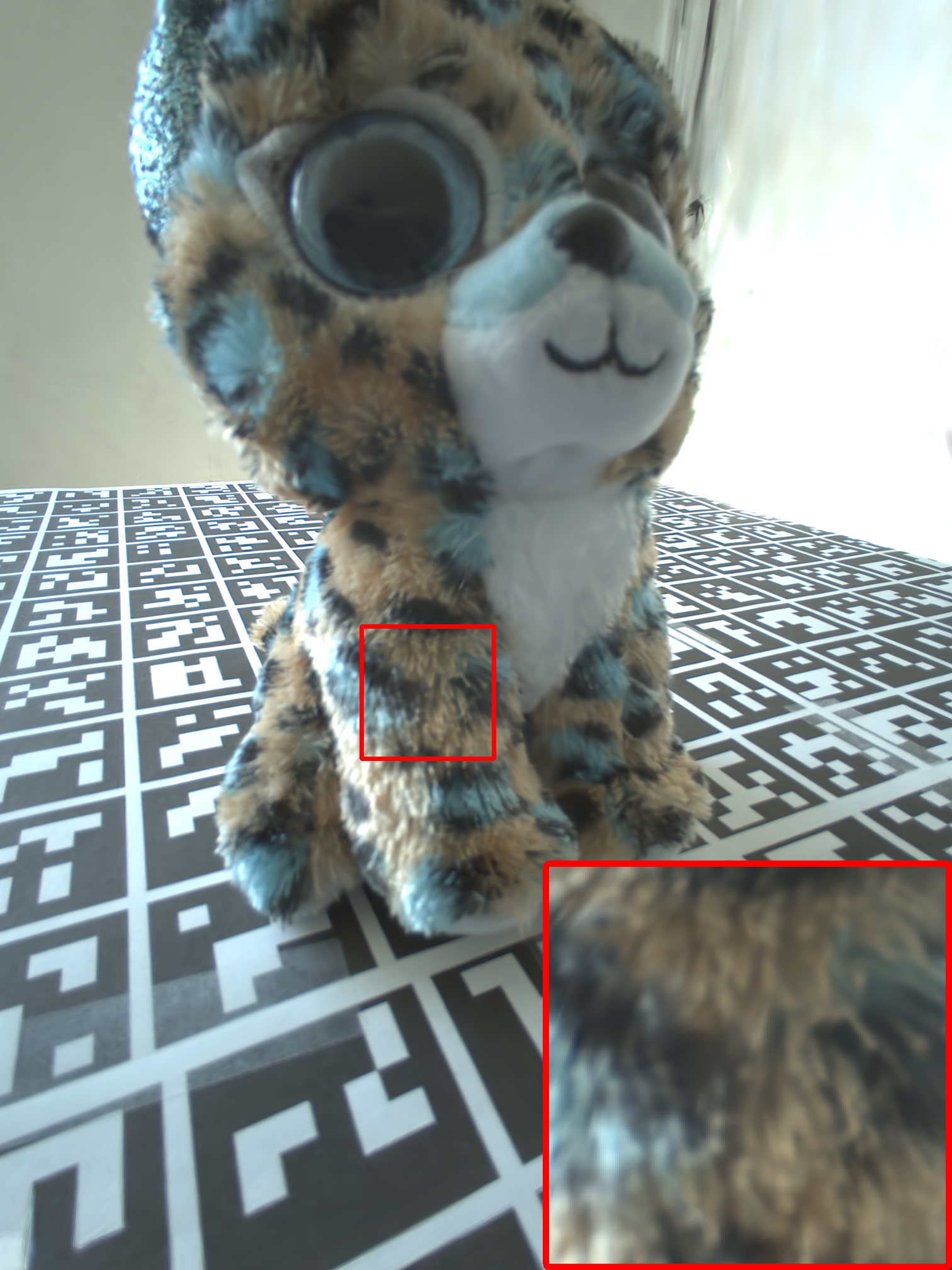}
    \caption{\textbf{Improved performance in 3D digital assets reconstruction.} Using Gaussian Splatting~\cite{kerbl20233d}, we reconstruct a small object with either captured photos (left) or our deblurred results (right) as inputs. A novel view is rendered as above for evaluation.}
    \label{fig:3d_recon}
\end{figure}

Smart glasses have revolutionized the way we interact with our surroundings, and 3D digital assets are one of the most exciting applications of this technology. By leveraging the camera and sensor capabilities of smart glasses, users can create detailed 3D digital assets of their interest. Reconstructing 3D objects at close distances is, in particular, important, as it can capture detailed textures and features for high-quality reconstruction. Our approach significantly improves the accuracy and performance of 3D reconstruction (Fig.~\ref{fig:3d_recon} and the supplementary video).

\section{Conclusion}  

In this paper, we introduce an efficient depth-varying dataset synthesis pipeline for defocus deblur and spatially-varying optical aberration correction in computational photography. Our method effectively bridges the gap between synthetic and real-world data by incorporating depth-varying defocus into spatially varying simulations, thereby providing a scalable and robust solution without requiring extensive real-world data collection. Experimental results demonstrate the superiority of our approach over alternative methods, both in terms of image restoration quality and in simulation speed and memory efficiency. This advancement establishes a strong foundation for future research in the field and significantly shortens the development cycle for computational photography algorithms.
{
    \small
    \bibliographystyle{ieeenat_fullname}
    \bibliography{references}
}

\end{document}